\documentclass[letterpaper]{article} 
\usepackage{aaai2027}  
\usepackage[hyphens]{url}  
\usepackage{graphicx} 
\usepackage{natbib}  
\usepackage{caption} 
\usepackage{booktabs}
\usepackage{amsmath}
\usepackage{amssymb}
\usepackage{amsfonts}
\usepackage{multirow}
\usepackage{subcaption}
\usepackage[table]{xcolor}
\usepackage{xspace}
\usepackage[most]{tcolorbox}
\usepackage{adjustbox}
\usepackage{dblfloatfix}
\definecolor{ourgray}{RGB}{235,235,235}
\definecolor{teacherblue}{RGB}{210,230,255}

\newtcolorbox{promptbox}[2][]{%
    enhanced,
    breakable,
    colback=white,
    colframe=gray!30!black,
    colbacktitle=gray!40!black,
    coltitle=white,
    fonttitle=\bfseries\sffamily\small,
    title={#2},
    sharp corners,
    boxrule=1.0pt,
    left=6pt, right=6pt, top=6pt, bottom=6pt,
    #1
}

\newcommand{\ours}{Flow3D-OPD\xspace}

\title{Flow3D-OPD: Multi-Teacher On-Policy Distillation for 3D Geometry Generation with Flow-Matching Diffusion Transformer}
\author{
    Zhiwei Ning\textsuperscript{1,4}\equalcontrib,
    Zhen Zhou\textsuperscript{4}\equalcontrib,
    Puhua Jiang\textsuperscript{4},
    Xintong Han\textsuperscript{4},
    Gengming Zhang\textsuperscript{1},
    Jie Yang\textsuperscript{1},\\
    Zhonglong Zheng\textsuperscript{2},
    Yuanjie Zheng\textsuperscript{3},
    Wei Liu\textsuperscript{1},
    Chunchao Guo\textsuperscript{4}
}
\affiliations{
    \textsuperscript{1}Shanghai Jiao Tong University
    \textsuperscript{2} Zhejiang Normal University
    \textsuperscript{3} Shandong Normal University
    \textsuperscript{4}Tencent Hunyuan3D \\
}

\begin{document}

\maketitle

\begin{abstract}
Recent image-to-3D generation models built on flow-matching diffusion Transformers (DiT) can produce high-fidelity meshes, yet their post-training strategy remains largely unexplored. There exist several critical bottlenecks in reinforcement learning: the inherent difficulty of defining comprehensive rewards for 3D geometric quality, and the gradient interference that arises when jointly optimizing heterogeneous objectives. Inspired by the practicability of on-policy distillation (OPD) in large language models and image generation, we propose \textbf{Flow3D-OPD}, a two-stage post-training framework that introduces multi-teacher distillation into 3D geometry generation. In the first stage, we utilize the semi-policy to enhance the foundational capability of the pretrained model and then design an agentic verifier for 3D geometric quality evaluation. Based on the verifier, we could cultivate domain-specialized teacher models via direct preference optimization (DPO). In the second stage, we consolidate heterogeneous expertise into a unified student model through on-policy distillation with hard task-routing sampling and gradient accumulation, which could mitigate the gradient interference in joint optimization. Without relying on elaborate modifications, our straightforward yet effective design achieves consistent improvements across all geometric quality dimensions and surpasses all teacher models in the average metric. Extensive experiments demonstrate that our approach provides an effective paradigm for reinforcement learning in 3D generation.
\end{abstract}

\section{Introduction}
\label{sec:intro}

\begin{figure}[!t]
    \centering
    \includegraphics[width=\columnwidth]{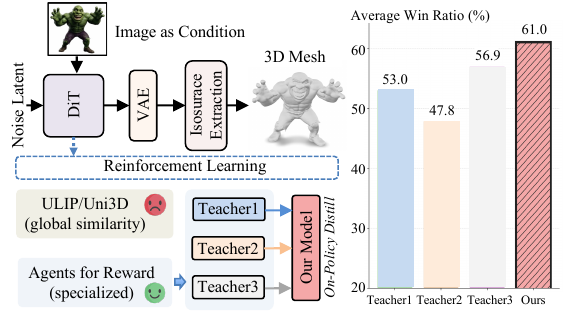}
    \caption{Our framework focuses on the on-policy distillation of the DiT module in the 3D generation pipeline. Our \ours enables to achieve a higher average win ratio than three specialized teachers.}
    \label{fig:intro}
\end{figure}

Recent 3D geometry generation pipelines ~\cite{li2025triposg, zhao2025hunyuan3d,lai2026lattice,lai2025hunyuan3d} employ a mixture architecture where a diffusion Transformer (DiT) ~\cite{peebles2023scalable} generates latent representations, and then the subsequent variational autoencoder (VAE) and isosurface extraction algorithm are utilized to decode the representation into explicit 3D meshes, as shown in Fig.~\ref{fig:intro}. The pipeline produces high-quality 3D assets from the image, enabling a wide range of applications in virtual reality, digital twins, and embodied AI~\cite{mildenhall2021nerf,poole2022dreamfusion}.

Despite this rapid advancement, post-training approaches through reinforcement learning (RL) remain absent from the 3D generation literature, which has long been employed in large language models~\cite{ouyang2022training,rafailov2023direct} and image generation~\cite{black2024training,liu2025flow}. The fundamental obstacle lies in the ambiguity of quality evaluation for 3D geometry. Unlike text-to-image generation where verifiers have been rapidly developed and proven effective~\cite{huang2023t2i,ghosh2023geneval}, precise quality assessment for 3D meshes requires multi-view rendering and geometric analysis, which relies on specific multi-modal reasoning. Existing approaches ~\cite{lai2026lattice,li2025triposg} utilize ULIP or Uni3D models ~\cite{xue2023ulip,zhou2024uni3d} to calculate global similarity between 2D images and 3D points, yet such a score cannot reflect fine-grained geometric dimensions such as orientation alignment, pose accuracy, and textural consistency. Moreover, jointly optimizing these heterogeneous objectives via diverse rewards will encounter gradient interference where updates favoring one dimension degrade others, which hinders consistent improvement in various dimensions.

However, recent image generation works introduce the on-policy distillation (OPD) strategy to improve model performance across multiple evaluation dimensions simultaneously. DiffusionOPD~\cite{li2026diffusionopd} and Flow-OPD~\cite{fang2026flow} prove that multi-teacher OPD can effectively unify various specialized capabilities into a single model by replacing sparse scalar rewards with dense velocity-field supervision along the denoising trajectory in the student model. This distillation paradigm could alleviate gradient interference in optimizing multiple dimensional quality in image generation. However, applying multi-teacher on-policy distillation to 3D geometry generation remains unexplored due to the difficulties in training domain-specialized teachers with 3D-specific rewards and consolidating these heterogeneous capabilities stably.

In this paper, we propose \ours, which innovatively introduces multi-teacher on-policy distillation into 3D geometry generation. Our approach follows a two-stage pipeline to overcome the above challenges. In the first stage, we train three teacher models targeting distinct evaluation dimensions. Specifically, we employ a semi-policy preference optimization strategy to enhance the base model's capability for fine-tuning, then decompose the quality evaluation into three different dimensions. Subsequently, an agentic verifier based on VLMs is designed for pairwise mesh comparison, which conducts multi-round structured reasoning with tool invocation and case memory mechanisms. Based on this verifier, we train the dedicated teacher for each quality dimension via direct preference optimization (DPO). In the second stage, we distill all three frozen teachers into a unified student model through on-policy distillation with hard task-routing and gradient accumulation mechanisms. At each training step, the student model generates its own denoising trajectory, and the corresponding teacher model provides step-level guidance along the hard task-routing. To alleviate gradient interference, we accumulate the gradient in a mini-batch for a single backward pass to ensure balanced multi-dimensional optimization. As shown on the right side of Fig.~\ref{fig:intro}, our student model successfully surpasses all teacher models on the average win-ratio metric. In summary, our contributions are three-fold as follows:
\begin{itemize}
    \item We propose the \ours approach, which is, to the best of our knowledge, the first to introduce on-policy distillation to 3D geometric generation through a two-stage training pipeline.
    \item We design an agentic reward system to enable DPO training of domain-specific teachers, followed by a multi-teacher distillation strategy with hard task-routing and gradient accumulation that consolidates their capabilities into a unified student model.
    \item  Extensive experiments demonstrate that our method consistently improves generation quality across all dimensions and achieves competitive or superior performance across multiple metrics.
\end{itemize}

\section{Related Work}
\label{sec:related}

\paragraph{3D Generation Models}
The field of 3D geometry generation has undergone several paradigm shifts. Primitive explorations leverage GANs~\cite{wu2016learning}, CLIP-guided synthesis~\cite{sanghi2022clip}, and Transformer-based approaches~\cite{yan2022shapeformer,yin2025shapegpt} to produce category-specific shapes. The subsequent works ~\cite{ho2020denoising,rombach2022high,poole2022dreamfusion,lin2023magic3d} are then introduced to enable text-to-3D and image-to-3D generation by leveraging pretrained 2D diffusion priors, though these methods require expensive per-instance optimization. A more recent line of work trains diffusion models natively on 3D latent spaces, achieving substantial quality improvements. Representative methods such as TripoSG~\cite{li2025triposg}, Hunyuan3D~\cite{zhao2025hunyuan3d,lai2025hunyuan3d}, and LATTICE~\cite{lai2026lattice} employ flow-matching on structured latent spaces to generate high-fidelity meshes. These systems usually follow a two-stage pipeline where the flow-matching generator produces latent point embeddings conditioned on the input image, followed by a variational autoencoder that decodes into signed distance fields for mesh extraction via differentiable marching cubes. Our work builds upon this architecture, focusing on the post-training alignment strategy that aims to improve generation quality beyond what pretraining alone can achieve.

\paragraph{Reinforcement Learning for Generative Models}
RL-based post-training has become an indispensable paradigm for further enhancing model capabilities~\cite{ouyang2022training}, progressing from PPO~\cite{schulman2017proximal} and DPO~\cite{rafailov2023direct} to GRPO~\cite{shao2024deepseekmath}. Its application has also been expanded from large language models to image generation. DDPO~\cite{black2024training} first demonstrates policy gradient optimization for diffusion models by treating each denoising step as an action in a Markov decision process. DPOK~\cite{fan2023dpok} introduces Kullback-Leibler (KL)-regularized optimization to prevent reward hacking, and DiffusionDPO~\cite{clark2024directly} adapts preference optimization to the denoising trajectory. In the flow-matching regime, Flow-GRPO~\cite{liu2025flow} leverages the SDE interpretation to define proper action distributions amenable to group-relative advantage estimation, while DiffusionNFT~\cite{zheng2025diffusionnft} proposes negative-free training that eliminates the need for a reference model. Despite these advances in 2D generation, RL-based optimization for 3D remains largely unexplored, primarily due to the difficulty of designing reliable multi-dimensional quality judgements for 3D meshes during training.

\begin{figure*}[!ht]
    \centering
    \includegraphics[width=0.9\textwidth]{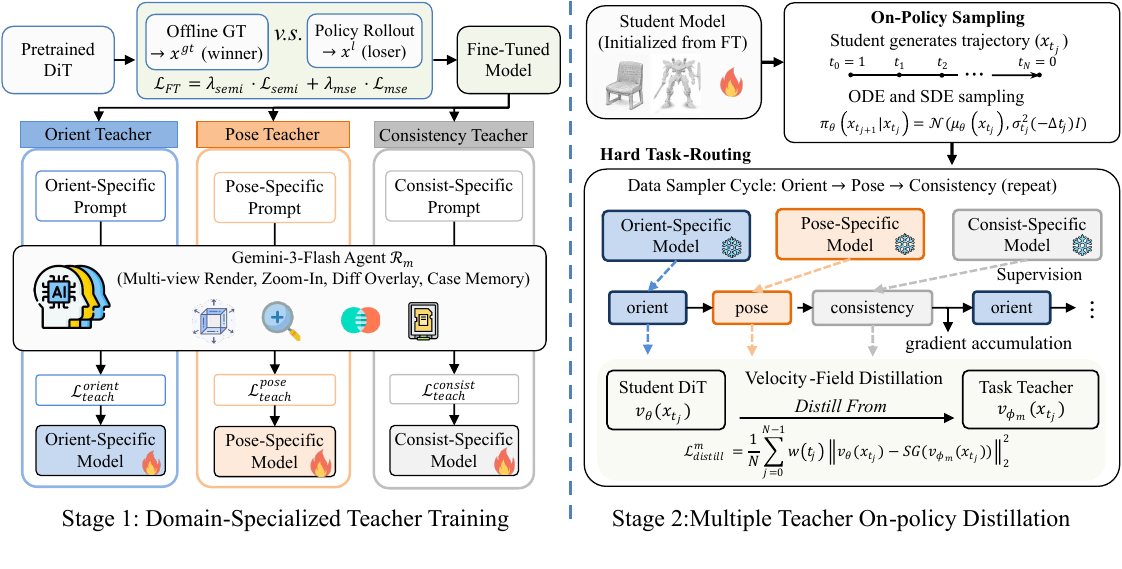}
    \vspace{-0.5em}
    \caption{Overview of our \ours framework. Stage 1 (left): the pretrained flow-matching DiT is first fine-tuned with $\mathcal{L}_\text{FT}$, then three teachers are trained through DPO with dimension-specific agentic reward. Stage 2 (right): the student model generates its own denoising trajectory via on-policy sampling, and a hard task-routing cycles through the three frozen teachers with mini-batch gradient accumulation.}
    \label{fig:framework}
    \vspace{-1.0em}
\end{figure*}

\paragraph{On-Policy Distillation}
On-policy distillation (OPD) addresses the distribution shift of offline distillation by requiring the student to generate trajectories from its own current policy under real-time teacher supervision~\cite{agarwal2024policy,gu2024minillm}. In the LLM, GKD~\cite{agarwal2024policy} first formalizes on-policy distillation by training on self-generated sequences to mitigate exposure bias. Subsequent work MiniLLM~\cite{gu2024minillm} adopts reverse KL to encourage the student to concentrate probability mass on high-quality modes, while DistiLLM~\cite{ko2024distillm} proposes skewed KL interpolation to balance mode-seeking and mode-covering behaviors. In visual generation, consistency distillation~\cite{song2023consistency,luo2023latent} first explores distilling multi-step diffusion models into fewer-step students. DiffusionOPD~\cite{li2026diffusionopd} extends OPD to diffusion models by exploiting the Markov structure of the denoising chain, deriving closed-form per-step KL divergence between Gaussian transition kernels. Flow-OPD~\cite{fang2026flow} further generalizes this to flow-matching architectures with stochastic differential equation (SDE) formulations and time-adaptive weighting, showing that multi-teacher distillation resolves the reward sparsity and gradient interference of joint multi-objective RL. Our work adapts the OPD mechanism on 3D geometry generation, where the inherently heterogeneous nature of quality assessment makes multi-teacher distillation a particularly natural fit.

\section{Preliminary}
\label{sec:preliminary}
\paragraph{DPO in Diffusion Models.}
Direct preference optimization (DPO)~\cite{rafailov2023direct} enables learning from pairwise preferences without explicit reward modeling. For language models, given winning and losing responses $x^w, x^l$ conditioned on prompt $c$, DPO directly optimizes the policy model $p_\theta$ via:
\begin{equation}                              
  \label{eq:dpo-llm}   
  \adjustbox{max width=0.9\linewidth}{$\displaystyle
  \mathcal{L}_{\text{DPO}}^{\text{LLM}}  = -\mathbb{E}\left[\log\sigma\left(\beta\log \frac{p_\theta(x^w|c)}{p_\text{ref}(x^w|c)} - \beta\log                          
  \frac{p_\theta(x^l|c)}{p_\text{ref}(x^l|c)}\right)\right],  
  $}
\end{equation}                                                                
where $\sigma$ is the sigmoid function, $\beta$ defines the penalty coefficient, and $p_\text{ref}$ indicates the reference model. DiffusionDPO~\cite{clark2024directly} extends this to diffusion models by deriving an equivalent loss over the denoising trajectory. Under the evidence lower bound (ELBO), the log-likelihood ratio decomposes into per-step noise prediction differences. The final loss for flow-matching models takes the form:
{\small
\begin{align}
\label{eq:dpo-diff}
\mathcal{L}_{\text{DPO}}^{\text{FM}}(\theta;x^w,x^l) &= -\mathbb{E}_{t}\left[\log\sigma\left(-\beta \omega(t) \cdot \Delta(\theta,t,x^w,x^l)\right)\right], \notag \\
\Delta(\theta, t,x^w,x^l) &= \left(\|v_\theta(x_t^w) - v^w\|_2^2 - \|v_\text{ref}(x_t^w) - v^w\|_2^2\right) \notag \\
&\quad - \left(\|v_\theta(x_t^l) - v^l\|_2^2 - \|v_\text{ref}(x_t^l) - v^l\|_2^2\right),
\end{align}
}where $ x_t^w$ and $x_t^l$ are noisy intermediates of the winner and loser at timestep $t$, defined via $x^{\{w,l\}}_t = (1-t)x^{\{w,l\}}+t\epsilon$. $v^w = \epsilon - x^w$ and $v^l = \epsilon - x^l$ are the target velocities. $v_\theta$ and $v_\text{ref}$ are the policy and reference velocity predictions, and $\omega(t)$ is a weighting function~\cite{ho2020denoising,kingma2021variational}.

\paragraph{On-Policy Distillation.}
On-policy distillation (OPD) addresses the distribution shift inherent in offline distillation by requiring the student to generate trajectories from its own current policy under real-time teacher supervision ~\cite{agarwal2024policy} . For autoregressive models, this is formulated as minimizing the reverse Kullback-Leibler (KL) divergence:
\begin{equation}
\footnotesize
\label{eq:opd-pre}
\mathcal{L}_{\text{OPD}} = -\mathbb{E}_{y \sim \pi_\theta} \left[ \log \frac{\pi_{\text{teacher}}(y|x)}{\pi_\theta(y|x)} \right] = D_{\text{KL}}(\pi_\theta \| \pi_{\text{teacher}}),
\end{equation}
where $\pi_\theta$ denotes the student policy, $\pi_{\text{teacher}}$ is the frozen teacher, and $y$ are trajectories sampled on-policy from the student. By training on self-generated distributions, OPD suppresses exposure bias and ensures the student receives supervision at the states it actually visits during inference.

\section{Our Method}
\label{sec:method}

Our model follows the architecture of Hunyuan3D-2.1~\cite{lai2025hunyuan3d}, where the flow-matching DiT generates structured latent $x \in \mathbb{R}^{4096 \times 64}$ conditioned on the image features. We employ a two-stage post-training strategy to optimize the pretrained DiT model, as illustrated in Fig.~\ref{fig:framework}.

\subsection{Stage 1: Domain-Specialized Teacher Training}
\label{sec:teacher-training}

Considering the defective capability of the basic model in rolling out high-quality samples for reliable preference learning, we first fine-tune the basic model via semi-policy preference optimization, and then train our teacher models on top of this fine-tuned (FT) model in stage 1.

\subsubsection{Fine-Tuning with Semi-Policy Preference Optimization.}
\label{sec:semi-policy}

In this phase, offline ground-truth 3D meshes serve as the preferred samples, while the current policy generates the rejected samples. Specifically, we replace $x^w$ in $\Delta(\theta, t,x^w,x^l)$ (Eq.~\ref{eq:dpo-diff}) with the ground-truth latent $x^\text{gt}$ to obtain the semi-policy DPO loss $\mathcal{L}_\text{semi}$. This formulation enables the model to optimize its own denoising trajectory along the stochastic differential equation (SDE) or ordinary differential equation (ODE) sampling. The overall fine-tuning loss combines $\mathcal{L}_\text{semi}$ with an mean squared error (MSE) loss function:
\begin{equation}
\mathcal{L}_\text{FT} = \lambda_\text{semi}\,\mathcal{L}_\text{semi} + \lambda_\text{mse}\,\mathbb{E}_{t}\!\left[\|v_\theta(x_t) - v^\text{gt}\|_2^2\right],
\end{equation}
where $\lambda_\text{semi}=0.2$, $\lambda_\text{mse}=0.8$. Based on the supervision, we enhance the foundational capability of the basic model and obtain the FT model for subsequent teacher model training.

\subsubsection{Agentic Reward Design and Teacher Training.}
\label{sec:reward}

We design an agentic reward system $\mathcal{R}_m$ based on Gemini-3-Flash~\cite{gemini2025flash} for pairwise mesh comparison. Given a 3D mesh pair $(\mathit{M}_a, \mathit{M}_b)$ and the reference image $\mathit{I}$, our agent performs multi-round structured reasoning:
\begin{equation}
(s_a, s_b) = \mathcal{R}_m(M_a, M_b, I; \mathcal{P}_m, \mathcal{T}, \mathcal{C}),
\end{equation}
where $\mathcal{P}_m$ is the dimension-specific evaluation prompt ($m \in \mathcal{M} =\{\text{orient}, \text{pose}, \text{consist}\}$), $\mathcal{T}$ denotes the tool set, including zoom-in rendering, viewpoint rotation, and difference-map overlay, which are necessary in geometric quality evaluation. And $\mathcal{C}$ is the case memory for judgment consistency. The agent is allowed up to four rounds of tool invocation before outputting structured preference scores $(s_a, s_b)$. Compared to global scalar metrics like ULIP/Uni3D, our agentic reward provides fine-grained and dimension-specific supervision with multi-view reasoning. Details of prompts and tools are provided in the supplementary material.

Building upon the agentic reward $\mathcal{R}_m$, we train each teacher $v_{\phi_m}$ via DPO. At each step, the sampling model rolls out two candidate latents $(x_a,x_b)$, which are then decoded to meshes and compared by the agentic reward to produce preference scores $(s_a, s_b)$. Based on the scores, we filter out pairs with a small margin $|s_a - s_b| < \delta$ and form the winner-loser pair $(x^w, x^l)$. Following Eq.~\ref{eq:dpo-diff}, each teacher is optimized via:
\begin{equation}
\label{eq:teacher-loss}
\mathcal{L}_{\text{teach}}^{m} = \mathcal{L}_{\text{DPO}}^{\text{FM}}(\theta_m; x^w,x^l), \quad m \in \mathcal{M} .
\end{equation}
In practice, we observe that the majority of sampled pairs yield score differences below $\delta$, leading to inefficient training under fully online rollout. To address this, we adopt an iterative ``rollout-then-train'' strategy: the model first generates a batch of candidate pairs over multiple steps and saves the valid ones filtered by $\delta$, then trains on this collected data before the next rollout stage. Based on these training objectives and strategies, we obtain three domain-specialized teacher models, each excelling at its target quality dimension.


\subsection{Stage 2: Multi-Teacher On-Policy Distillation}
\label{sec:opd}

Given three trained teacher models $\{v_{\phi_m}\}_{m\in \mathcal{M}}$, we then integrate them into a single student $v_\theta$ with on-policy distillation supervision in this stage.

\subsubsection{On-Policy Distillation in Flow-Matching.}

Following the OPD loss function formulated in Eq.~\ref{eq:opd-pre}, we minimize the reverse KL divergence between the student and teacher transition kernels at each step along the student's on-policy trajectory. Under the SDE discretization (Euler-Maruyama), the one-step transition kernel is Gaussian:
\begin{equation}
\label{eq:transition-kernel}
\vspace{-0.2em}
\pi_\theta(x_{t_{j+1}} | x_{t_j}) = \mathcal{N}\big(\mu_\theta(x_{t_j}),\; \sigma_{t_j}^2(-\Delta t_j) \mathit{I}\big),
\end{equation}
where $1=t_0>t_1>\cdots t_N=0$ define the denoising timesteps and $\Delta t_j = t_{j+1}-t_j<0$. $\sigma_{t_j} $ is the per-step diffusion coefficient determined solely by the noise schedule, and the transition mean $\mu_\theta$ is deterministically parameterized by the student velocity field. The teacher transition kernel $\pi_{\phi_m}$ uses the teacher velocity to compute its mean $\mu_{\phi_m}$ but shares the same covariance $\bar{\sigma}_{t_j}^2=\sigma_{t_j}^2(-\Delta t_j)$. Since both kernels share the same isotropic covariance, their reverse KL admits a closed form:
\begin{align}
\label{eq:kl-closed}
\vspace{-0.5em}
\text{KL}\big(\pi_\theta \| \pi_{\phi_m}\big) &= \text{KL}\big(\mathcal{N}\big(\mu_\theta,\; \bar{\sigma}_{t_j}^2 \mathit{I}\big) \| \mathcal{N}\big(\mu_{\phi_m},\; \bar{\sigma}_{t_j}^2 \mathit{I}\big)\big) \notag \\
&= \frac{\|\mu_\theta(x_{t_j}) - \mu_{\phi_m}(x_{t_j})\|_2^2}{2\bar{\sigma}_{t_j}^2}.
\end{align}
The stochastic transition degenerates to a deterministic map under the ODE regime, and distribution matching reduces to direct velocity-field alignment with L2 loss. The reverse KL defines the per-step distillation objective for a single teacher.

\subsubsection{Hard Task-Routing and Mini-Batch Gradient Accumulation.}

To consolidate all three teachers while mitigating gradient interference, we introduce a hard task-routing mechanism. At each training step, the sample carries a data-type label $m \in \{\text{orient}, \text{pose}, \text{consist}\}$ that deterministically routes supervision to the corresponding teacher $v_{\phi_m}$, so that its gradient stays free from cross-teacher conflict. The routed OPD loss aggregates per-step velocity-matching objectives along the on-policy trajectory via:
\begin{flalign}
  \label{eq:opd-loss}
  \vspace{-1.5em}
  &\adjustbox{max width=0.9\linewidth}{$\displaystyle
  \mathcal{L}^{m}_\text{distill}(\theta) =
  \frac{1}{N}\sum_{j=0}^{N-1} w(t_j)
  \big\| v_\theta(x_{t_j}) -\mathrm{SG}( v_{\phi_m}(x_{t_j}) )\big\|_2^2
  $} 
  \end{flalign}
where the stop-gradient operator $\mathrm{SG}(\cdot)$ detaches the teacher output. The weighting function $w(t_j)$ controls the distillation strength across different denoising timestep $t_j$, which equals $\frac{\Delta t_j}{2} \left( \frac{\sigma_{t_j}(1-t_j)}{2t_j} + \frac{1}{\sigma_{t_j}} \right)^2$ under SDE sampling and reduces to a constant $1$ under ODE sampling. The detailed derivation is provided in the supplementary material.

\begin{table*}[ht]
\centering
\resizebox{1.0\textwidth}{!}{
\begin{tabular}{lcccccc|cc|cc|c}
\toprule
\multirow{3}{*}{Method} & \multicolumn{8}{c|}{Gemini-3-Flash} & \multicolumn{2}{c|}{GPT-5.5} & \multirow{3}{*}{User Study $\uparrow$} \\
& \multicolumn{2}{c}{Orient-Eval} & \multicolumn{2}{c}{Pose-Eval} & \multicolumn{2}{c|}{Consistency-Eval} & \multicolumn{2}{c|}{Average} & \multicolumn{2}{c|}{Average} &  \\
 & A wins $\downarrow$ & B wins $\uparrow$ & A wins $\downarrow$ & B wins $\uparrow$ & A wins $\downarrow$ & B wins $\uparrow$ & A wins $\downarrow$ & B wins $\uparrow$ & A wins $\downarrow$ & B wins $\uparrow$ & \\
\midrule
FT (semi-policy) & 5.5 & 11.2 & 0.5 & 85.1 & 12.4 & 79.4 & 5.9 & 54.9 & 15.7 & 45.7 & -- \\
\midrule
Orient-Teacher & \cellcolor{ourgray}\textbf{2.7} & \cellcolor{ourgray}\textbf{22.7} & 0.8 & 89.0 & 30.8 & 53.2 & 10.3 & 53.0 & 13.4 & 47.3 & 4.14 \\
Pose-Teacher & 10.1 & 14.3 & \cellcolor{ourgray}\textbf{0.0} & \cellcolor{ourgray}\textbf{93.7} & 41.7 & 41.7 & 16.0 & 47.9 & 15.1 & 43.6 & 4.06 \\
Consistency-Teacher & 6.5 & 15.9 & 1.0 & 83.5 & \cellcolor{ourgray}\textbf{8.0} & \cellcolor{ourgray}\textbf{82.0} & \underline{5.2} & 56.9 & 11.8 & \underline{50.9} & 3.40 \\
\midrule
DPO & 13.4 & 7.2 & 19.1 & 71.8 & 13.0 & 77.0 & 15.1 & 48.3 & 20.4 & 41.5 & 3.56 \\
FT + DPO & 5.9 & 14.4 & 0.9 & 91.8 & 14.0 & 76.0 & 6.6 & 57.2 & 14.1 & 44.7 & 4.08 \\
OPD & 10.9 & 10.9 & \textbf{0.0} & 93.4 & 15.0 & 79.0 & 8.6 & \underline{57.3} & \underline{11.6} & 48.3 & \underline{4.36} \\
\rowcolor{ourgray} FT + OPD (Ours) & \underline{5.1} & \underline{18.8} & \textbf{0.0} & \underline{93.5} & \underline{10.0} & \underline{81.0} & \textbf{4.9} & \textbf{61.0} & \textbf{9.9} & \textbf{51.4} & \textbf{4.40} \\
\bottomrule
\end{tabular}
}
\caption{Pairwise evaluation across three quality dimensions, with cross-evaluator and human validation. Model A is the pretrained baseline (Hunyuan3D-2.1); Model B is the evaluated model. A/B wins (\%) denote the percentage of cases each is judged superior. The left columns use Gemini-3-Flash; GPT-5.5 Avg reports averaged results under the independent GPT-5.5 evaluator; User Study reports the average human ranking score. Best results are in \textbf{bold}, second best are \underline{underlined}.}
\label{tab:gemini-eval}
\end{table*}

\begin{table}[ht]
\centering
\resizebox{\columnwidth}{!}{
\begin{tabular}{l|cccc}
\toprule
Method & ULIP-T & ULIP-I & Uni3D-T  & Uni3D-I  \\
\midrule
Craftsman 1.5 & 0.074 & 0.129 & 0.233& 0.299\\
Michelangelo & 0.075 & 0.115 & 0.206& 0.261 \\
Direct3D-s2 & 0.074 & 0.122 & 0.247 & 0.314 \\
Hi3DGen & 0.076 & 0.131 & 0.237 & 0.299 \\
Trellis2 & 0.080 & 0.129 & 0.241 & 0.307 \\
HY3D 2.0 & 0.077 & 0.130 & 0.245 & 0.315 \\
\midrule
HY3D 2.1 & 0.081 & 0.133& 0.246 & 0.314\\
\rowcolor{ourgray}\ours(Ours) & \textbf{0.083} & \textbf{0.138} & \textbf{0.252} & \textbf{0.321}\\
\bottomrule
\end{tabular}
}
\caption{Comparison with previous works on 3D--text (T) and 3D--image (I) similarity calculated by ULIP/Uni3D models. The best results are in \textbf{bold}.}
\label{tab:benchmark}
\end{table}

While hard task-routing keeps each mini-batch conflict-free, a single update from one dimension alone is still biased toward that teacher. We therefore organize training data such that every three consecutive mini-batches cycle through all data types in round-robin order: orient $\to$ pose $\to$ consistency. Gradient accumulation across this full cycle before each optimizer step ensures every parameter update incorporates guidance from all three teachers simultaneously:
\begin{equation}
\label{eq:grad-accum}
\mathcal{L}^{\text{total}}_\text{distill} = \frac{1}{|\mathcal{M}|}\sum_{m\in\mathcal{M}} \mathcal{L}_\text{distill}^{m}(\theta).
\end{equation}
In this way, hard task-routing isolates each teacher's supervision at the mini-batch level, while gradient accumulation aggregates them into a balanced consensus at the update level, substantially mitigating the gradient interference of multi-objective optimization.

\section{Experiments}
\label{sec:experiments}

\subsection{Experimental Setup}

\paragraph{Training Settings.} We build upon Hunyuan3D-2.1~\cite{lai2025hunyuan3d}, a 3B-parameter DiT architecture for 3D latent representation generation. Three domain-specialized teachers are refined from the fine-tuned model via on-policy DPO with learning rate $1\times10^{-6}$, SDE sampling with noise level $a=0.3$, and a minimum confidence margin of $\delta=0.167$ across $2 \times 8$ H20 GPUs. During the second stage, the student is also initialized from the fine-tuned model and distilled by randomly sampling states along the 50-step denoising trajectory. To handle the imbalanced sizes of the three sub-datasets, we employ dataset sampling with replacement across data types. At inference, we apply classifier-free guidance with a default scale of $3.0$.

\paragraph{Evaluation Metrics.} We adopt two complementary evaluation protocols. For multi-dimensional quality assessment, we employ pairwise judgement to report the win rates across orientation, pose, and consistency dimensions. For overall generation quality evaluation, we report ULIP~\cite{xue2023ulip} and Uni3D~\cite{zhou2024uni3d} scores to measure 3D--text and 3D--image cosine similarities in LATTICE-Bench ~\cite{lai2026lattice}. We also provide qualitative visualizations of generated 3D meshes for intuitive comparison.

\subsection{Main Results}

\paragraph{Multi-Dimensional Quality Evaluation}

Table~\ref{tab:gemini-eval} demonstrates the effectiveness of our method across multiple quality dimensions. We compare the meshes generated by each model (denoted as model B) against those from the pretrained baseline Hunyuan3D-2.1~\cite{lai2025hunyuan3d} (denoted as model A) with our agentic verifier for pairwise evaluation. ``A wins'' indicates the ratio that the baseline produces higher quality, while ``B wins'' indicates that the evaluated model is superior. Compared with the fine-tuned (FT) model, each teacher achieves strong performance on its target dimension but degrades on others. For instance, the orient teacher achieves the lowest A wins (2.7\%) on orientation evaluation but causes consistency to drop significantly (30.8\% A wins). This phenomenon confirms that the three evaluated dimensions are inherently heterogeneous, and agentic-based DPO can yield substantial improvements on specific target objectives.

We further present our final \ours result and compare with different training strategies at the bottom of the table. Directly applying DPO on the initial model without fine-tuning leads to degraded performance due to insufficient basic capability. Applying OPD without the semi-policy stage yields moderate improvements across all dimensions but falls short of the full pipeline. In contrast, our \ours achieves the best overall balance: 18.8\%, 93.5\%, and 81\% B wins ratio on orient, pose, and consistency, respectively. Finally, our method consistently outperforms all teachers on the averaged metrics with 61.0\% B wins.

To validate the generalization and effectiveness of our model, we additionally conduct two complementary assessments on the right of Table~\ref{tab:gemini-eval}. First, we employ GPT-5.5 as an independent evaluator, which is a different model from the Gemini-3-Flash used in the teacher training reward. Our \ours model also achieves the lowest A wins (9.9\%) and highest B wins (51.4\%) under GPT-5.5, confirming that the observed improvements are not artifacts of evaluator-specific optimization. Second, we conduct a blind user study where human annotators rank all methods by their generation quality from best to worst without knowing model identities. Our method also achieves the highest average ranking score, further validating that the improvements reflect genuine 3D quality gains perceived by human evaluators.

\begin{table}[!t]
\centering
\resizebox{\columnwidth}{!}{
\begin{tabular}{lccc}
\toprule
Update Strategy & A-wins $\downarrow$ & B-wins$\uparrow$& Uni3D-I \\
\midrule
Single-step (sequential) & 7.1 & 59.8 & 0.319 \\
Single-step (hard task-routing) & 6.6 & 58.5 & 0.320 \\
\midrule
\rowcolor{ourgray} Multi-step (hard task-routing) & \textbf{4.9} & \textbf{61.0} & \textbf{0.321} \\
\bottomrule
\end{tabular}
}
\caption{Ablation on different gradient update strategies during OPD. Best results are in \textbf{bold}.}
\label{tab:ablation-update}
\end{table}

\begin{table}[!t]
\centering
\resizebox{1.0\columnwidth}{!}{
\begin{tabular}{c|ccccc}
\toprule
Stage & Strategy & A-wins $\downarrow$ & B-wins$\uparrow$& Uni3D-T & Uni3D-I \\
\midrule
\multirow{2}{*}{FT} & ODE & 6.4 & 51.7 & 0.245 & 0.314 \\
 & \cellcolor{ourgray}SDE & \cellcolor{ourgray}5.9 & \cellcolor{ourgray}54.9 & \cellcolor{ourgray}0.248 & \cellcolor{ourgray}0.315 \\
\midrule
\multirow{2}{*}{OPD} & \cellcolor{ourgray}ODE & \cellcolor{ourgray}\textbf{4.9} & \cellcolor{ourgray}\textbf{61.0} & \cellcolor{ourgray}\textbf{0.252} & \cellcolor{ourgray}\textbf{0.321} \\
 & SDE & 7.3 & 54.8 & 0.251 & 0.319  \\
\bottomrule
\end{tabular}
}
\caption{Ablation on SDE and ODE sampling in both FT and OPD stages (NFE=50). Best results are in \textbf{bold}.}
\label{tab:ablation-sde-ode}
\end{table}




\begin{figure}[!t]
    \centering
    \includegraphics[width=\columnwidth]{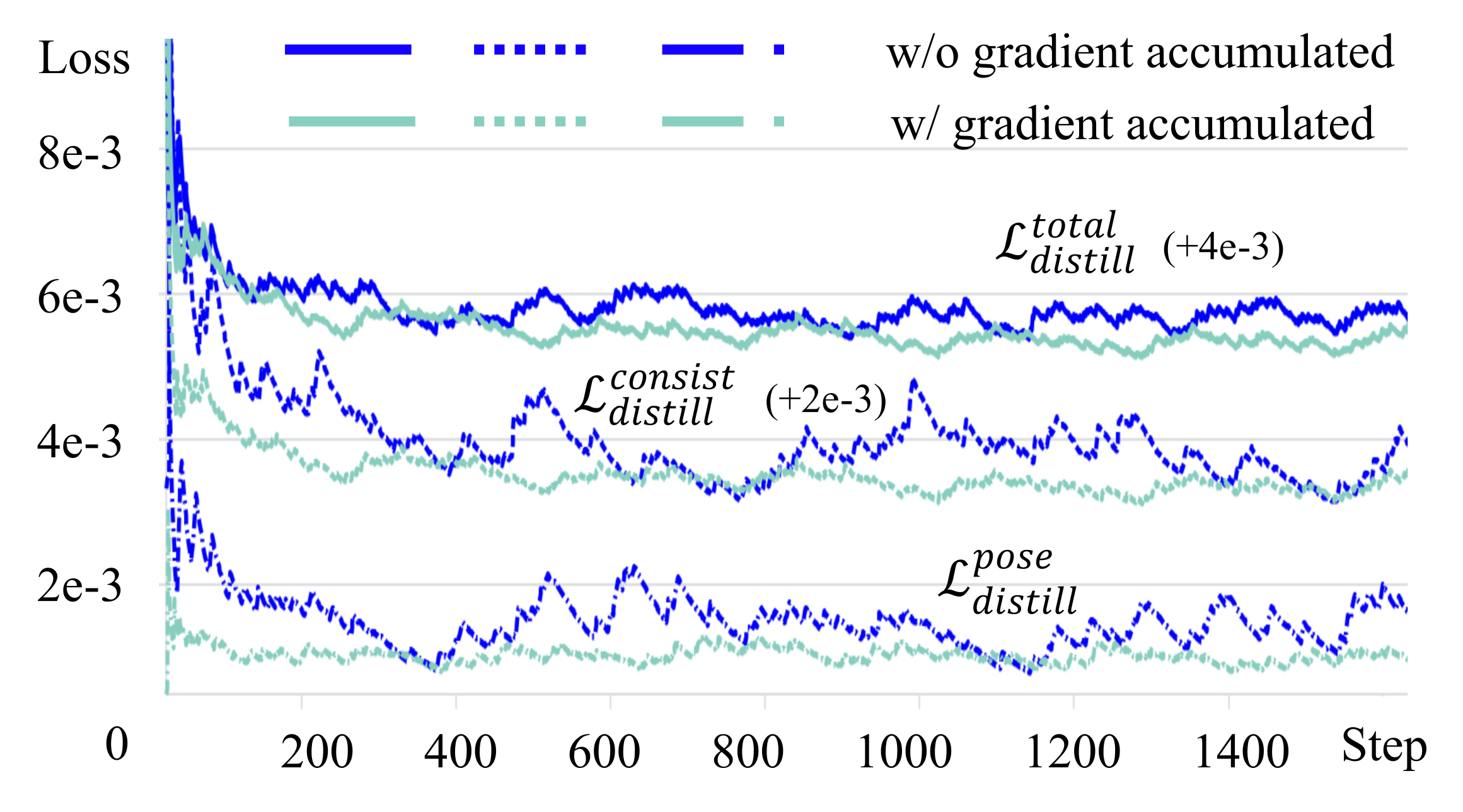}
    \caption{Distillation loss curves under different gradient update strategies. Gradient accumulation in mini-batch exhibits smoother convergence and lower final loss compared to single-step updates.}
    \label{fig:ablation-curves}
    \vspace{-1.0em}
\end{figure}

\paragraph{Comparison with Previous Methods}

Table~\ref{tab:benchmark} compares our method with previous 3D generation approaches on previous 3D--text and 3D--image similarity benchmarks~\cite{lai2026lattice}, including Craftsman 1.5~\cite{li2024craftsman3d}, Michelangelo~\cite{zhao2023michelangelo}, Trellis2/512~\cite{xiang2026native}, Direct3D-S2~\cite{wu2026direct3d}, Hi3DGen~\cite{ye2025hi3dgen}, and Hunyuan3D 2.0~\cite{zhao2025hunyuan3d}. We evaluate the global similarity with  ULIP~\cite{xue2023ulip} and Uni3D~\cite{zhou2024uni3d} metrics. Results demonstrate that our \ours achieves consistent gains over the baseline Hunyuan3D-2.1 ~\cite{lai2025hunyuan3d} across all four metrics and outperforms all compared previous methods.

\begin{figure}[!ht]
    \centering
    \includegraphics[width=\columnwidth]{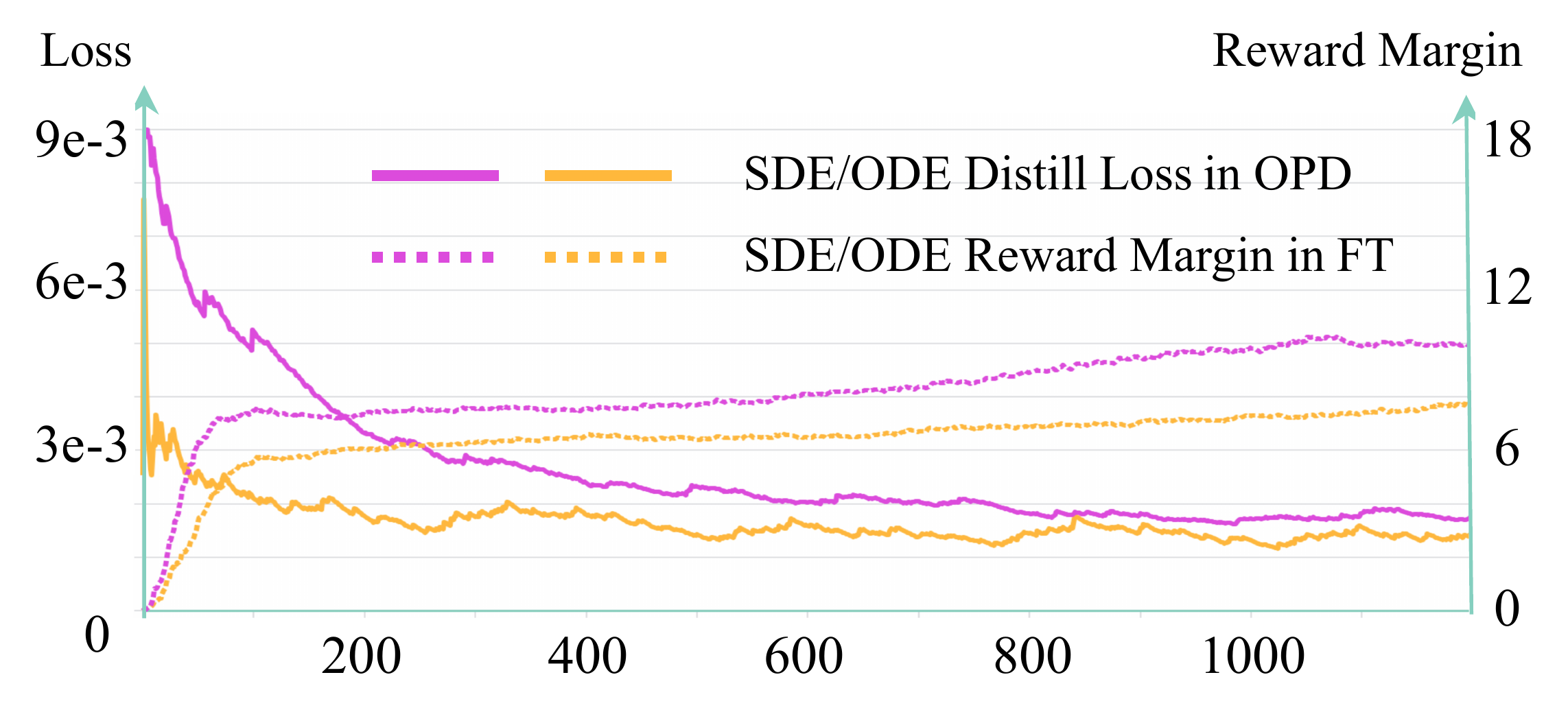}
    \caption{Reward margin and distill loss curves in SDE/ODE Sampling. SDE achieves higher reward margins during FT stage, indicating more discriminative preference pairs. ODE converges faster and reaches a lower final loss than SDE during the OPD stage.}
    \label{fig:ablation-sde-ode}
    \vspace{-1.0em}
\end{figure}

\subsection{Ablation Studies}

\paragraph{Effect of Gradient Update Strategy}
We compare three gradient update strategies during the OPD stage in Table~\ref{tab:ablation-update}. The first is single-step backward and uses sequential sampling, where all samples of one dimension are processed before moving to the next. The second is also a single-step backward strategy but with hard task-routing sampling. The third is our gradient accumulation strategy with hard task-routing sampling. The results suggest that our configuration achieves the best performance across all metrics, while hard task-routing sampling is better than sequential sampling. Moreover, we compare the distillation loss curves of single-step and multi-step updates under the same hard task-routing sampling strategy, as shown in Fig.~\ref{fig:ablation-curves}. Our gradient accumulation strategy produces more stable distillation loss curves than the single-step backward variant, and also converges to a lower final loss across both total and per-task objectives. This confirms that accumulating multi-task gradients in a mini-batch before each update effectively prevents inter-task interference and achieves better performance.

\begin{figure*}[!ht]
    \centering
    \includegraphics[width=\textwidth]{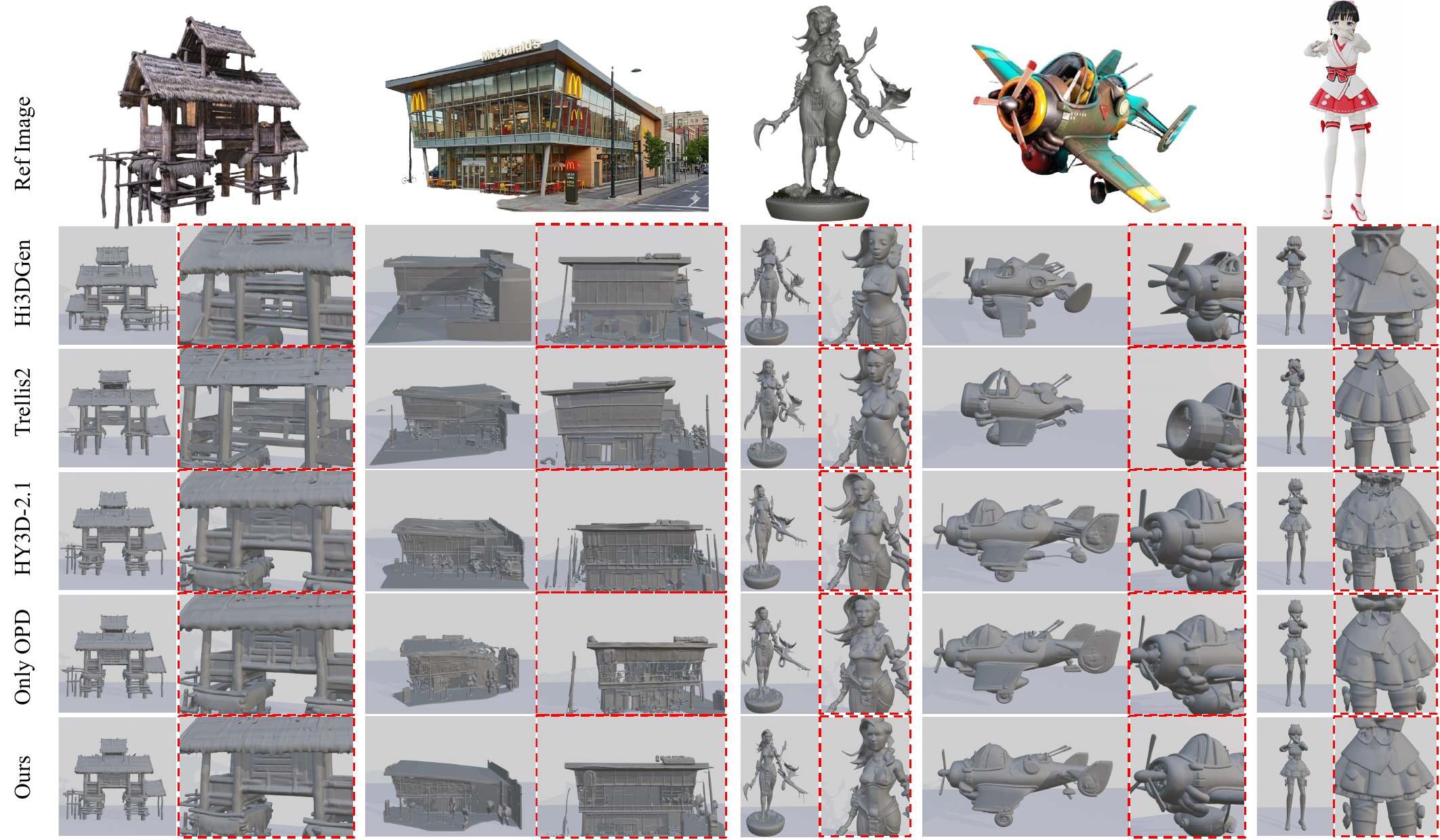}
    \caption{Qualitative comparison on 3D meshes generated with various approaches or training strategies. Our model produces 3D meshes with accurate and consistent structure.}
    \label{fig:qualitative}
\vspace{-1.0em}
\end{figure*}

\begin{figure}[!ht]
    \centering
    \includegraphics[width=\columnwidth]{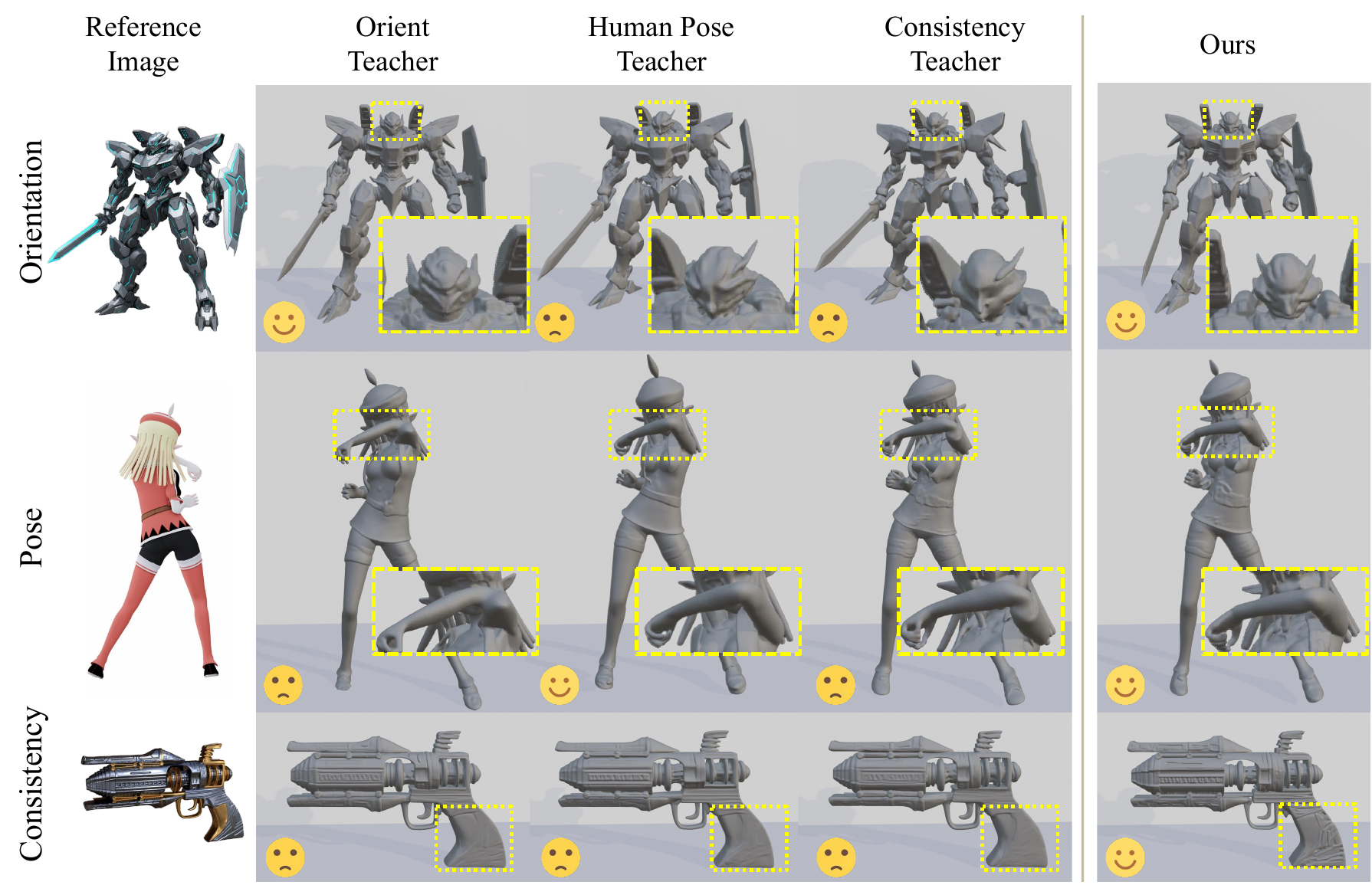}
    \caption{Visualization comparison between our model and individual teachers on orientation (top), pose (middle), and consistency (bottom) validation dimensions.}
    \label{fig:teacher-compare}
\vspace{-1.0em}
\end{figure}

\paragraph{Effect of SDE vs. ODE Sampling}
We compare the effectiveness of SDE and ODE sampling mechanisms across the FT and OPD stages in Table~\ref{tab:ablation-sde-ode}. In the FT stage, SDE sampling outperforms ODE across both global and specialized metrics. We attribute this to the additional stochasticity that generates more diverse on-policy losers and produces higher reward margins between winner and loser pairs, as demonstrated by the dashed curves in Fig.~\ref{fig:ablation-sde-ode}. Therefore, we initialize the student model with FT model with SDE sampling. However, in the OPD stage, the deterministic ODE formulation proves more effective than SDE. As illustrated in the bottom of Table~\ref{tab:ablation-sde-ode} and Fig.~\ref{fig:ablation-sde-ode}, ODE achieves better results with faster loss convergence and lower distillation loss. This is because the student model can align more precisely with the teacher's velocity field at each step without the interference introduced by the stochastic diffusion term. This asymmetric finding suggests that exploration diversity benefits preference learning during model fine-tuning, while deterministic trajectory is more suitable for distillation in the OPD stage.


\subsection{Qualitative Visualization}

Fig.~\ref{fig:qualitative} presents qualitative comparisons across various approaches or training strategies: Hi3DGen, Trellis2, Hunyuan3D-2,1, OPD-only (distillation without the FT stage), and our full pipeline \ours model. Previous models exhibit limited capability in generating fine-grained details, such as the geometric structures of buildings and the propellers of aircraft models. Our baseline model, Hunyuan3D-2.1, produces geometrically plausible but misaligned or structurally inconsistent meshes. OPD-only model yields moderate improvements yet lacks the stable foundation for consistent high-quality generation. Our full model produces meshes with coherent and precise multi-view structure, demonstrating the complementary benefits of basic capability scaling and multi-teacher on-policy distillation.

Fig.~\ref{fig:teacher-compare} further compares our distilled student model against the three individual teacher models on their respective areas of expertise. In the first row of orientation validation, the pose teacher and consistency teacher fail to align the mesh to the canonical front-facing direction, which further introduces asymmetric artifacts in the head region, while our student maintains correct orientation. In the second row of pose validation, the orient teacher and consistency teacher produce visible deformations at the elbows of the people, whereas our student preserves faithful articulated pose though it is unseen in the reference image. In the third row, our student exhibits richer detail in the surface texture compared to the non-specialized teachers. These results confirm that the student model unifies the strengths of all three teachers while avoiding their individual weaknesses in other dimensions.


\section{Conclusion}
\label{sec:conclusion}

We present \ours, a two-stage post-training framework that first introduces multi-teacher on-policy distillation into 3D geometry generation. In the first stage, we design an agentic reward system for fine-grained geometric evaluation and train domain-specialized teachers via DPO across three quality dimensions. In the second stage, multiple teacher distillation with hard task-routing and gradient accumulation enables the student to absorb heterogeneous expertise through velocity-field supervision along its own denoising trajectory. Extensive experiments demonstrate that \ours achieves consistent improvements across all geometric dimensions and surpasses teachers on the averaged metric. 


\bibliography{references}

@inproceedings{peebles2023scalable,
  title={Scalable diffusion models with transformers},
  author={Peebles, William and Xie, Saining},
  booktitle={Proceedings of the IEEE/CVF international conference on computer vision},
  pages={4172--4182},
  year={2023}
}

@article{li2025triposg,
  title={Triposg: High-fidelity 3d shape synthesis using large-scale rectified flow models},
  author={Li, Yangguang and Zou, Zi-Xin and Liu, Zexiang and Wang, Dehu and Liang, Yuan and Yu, Zhipeng and Liu, Xingchao and Guo, Yuan-Chen and Liang, Ding and Ouyang, Wanli and others},
  journal={IEEE Transactions on Pattern Analysis and Machine Intelligence},
  year={2025},
  publisher={IEEE}
}

@article{zhao2025hunyuan3d,
  title={Hunyuan3d 2.0: Scaling diffusion models for high resolution textured 3d assets generation},
  author={Zhao, Zibo and Lai, Zeqiang and Lin, Qingxiang and Zhao, Yunfei and Liu, Haolin and Yang, Shuhui and Feng, Yifei and Yang, Mingxin and Zhang, Sheng and Yang, Xianghui and others},
  journal={arXiv preprint arXiv:2501.12202},
  year={2025}
}

@article{lai2025hunyuan3d,
  title={Hunyuan3d 2.5: Towards high-fidelity 3d assets generation with ultimate details},
  author={Lai, Zeqiang and Zhao, Yunfei and Liu, Haolin and Zhao, Zibo and Lin, Qingxiang and Shi, Huiwen and Yang, Xianghui and Yang, Mingxin and Yang, Shuhui and Feng, Yifei and others},
  journal={arXiv preprint arXiv:2506.16504},
  year={2025}
}

@inproceedings{lai2026lattice,
  title={Lattice: Democratize high-fidelity 3d generation at scale},
  author={Lai, Zeqiang and Zhao, Yunfei and Zhao, Zibo and Liu, Haolin and Lin, Qingxiang and Huang, Jingwei and Guo, Chunchao and Yue, Xiangyu},
  booktitle={Proceedings of the IEEE/CVF Conference on Computer Vision and Pattern Recognition},
  pages={19982--19992},
  year={2026}
}

@article{li2026diffusionopd,
  title={DiffusionOPD: A unified perspective of on-policy distillation in diffusion models},
  author={Li, Quanhao and Yu, Junqiu and Jiang, Kaixun and Wei, Yujie and Xing, Zhen and Li, Pandeng and Chu, Ruihang and Zhang, Shiwei and Liu, Yu and Wu, Zuxuan},
  journal={arXiv preprint arXiv:2605.15055},
  year={2026}
}

@article{fang2026flow,
  title={Flow-OPD: On-policy distillation for flow matching models},
  author={Fang, Zhen and Huang, Wenxuan and Zeng, Yu and Zhao, Yiming and Chen, Shuang and Feng, Kaituo and Lin, Yunlong and Chen, Lin and Chen, Zehui and Cao, Shaosheng and others},
  journal={arXiv preprint arXiv:2605.08063},
  year={2026}
}

@article{liu2025flow,
  title={Flow-grpo: Training flow matching models via online rl},
  author={Liu, Jie and Liu, Gongye and Liang, Jiajun and Li, Yangguang and Liu, Jiaheng and Wang, Xintao and Wan, Pengfei and Zhang, Di and Ouyang, Wanli},
  journal={arXiv preprint arXiv:2505.05470},
  year={2025}
}

@article{zheng2025diffusionnft,
  title={Diffusionnft: Online diffusion reinforcement with forward process},
  author={Zheng, Kaiwen and Chen, Huayu and Ye, Haotian and Wang, Haoxiang and Zhang, Qinsheng and Jiang, Kai and Su, Hang and Ermon, Stefano and Zhu, Jun and Liu, Ming-Yu},
  journal={arXiv preprint arXiv:2509.16117},
  year={2025}
}

@article{rafailov2023direct,
  title={Direct preference optimization: Your language model is secretly a reward model},
  author={Rafailov, Rafael and Sharma, Archit and Mitchell, Eric and Manning, Christopher D and Ermon, Stefano and Finn, Chelsea},
  journal={Advances in neural information processing systems},
  volume={36},
  pages={53728--53741},
  year={2023}
}

@inproceedings{black2024training,
  title={Training diffusion models with reinforcement learning},
  author={Black, Kevin and Janner, Michael and Du, Yilun and Kostrikov, Ilya and Levine, Sergey},
  booktitle={International Conference on Learning Representations},
  volume={2024},
  pages={4965--4987},
  year={2024}
}

@article{ouyang2022training,
  title={Training language models to follow instructions with human feedback},
  author={Ouyang, Long and Wu, Jeffrey and Jiang, Xu and Almeida, Diogo and Wainwright, Carroll and Mishkin, Pamela and Zhang, Chong and Agarwal, Sandhini and Slama, Katarina and Ray, Alex and others},
  journal={Advances in neural information processing systems},
  volume={35},
  pages={27730--27744},
  year={2022}
}

@article{schulman2017proximal,
  title={Proximal policy optimization algorithms},
  author={Schulman, John and Wolski, Filip and Dhariwal, Prafulla and Radford, Alec and Klimov, Oleg},
  journal={arXiv preprint arXiv:1707.06347},
  year={2017}
}

@article{shao2024deepseekmath,
  title={Deepseekmath: Pushing the limits of mathematical reasoning in open language models},
  author={Shao, Zhihong and Wang, Peiyi and Zhu, Qihao and Xu, Runxin and Song, Junxiao and Bi, Xiao and Zhang, Haowei and Zhang, Mingchuan and Li, YK and Wu, Yang and others},
  journal={arXiv preprint arXiv:2402.03300},
  year={2024}
}

@article{fan2023dpok,
  title={Dpok: Reinforcement learning for fine-tuning text-to-image diffusion models},
  author={Fan, Ying and Watkins, Olivia and Du, Yuqing and Liu, Hao and Ryu, Moonkyung and Boutilier, Craig and Abbeel, Pieter and Ghavamzadeh, Mohammad and Lee, Kangwook and Lee, Kimin},
  journal={Advances in Neural Information Processing Systems},
  volume={36},
  pages={79858--79885},
  year={2023}
}

@inproceedings{clark2024directly,
  title={Directly fine-tuning diffusion models on differentiable rewards},
  author={Clark, Kevin and Vicol, Paul and Swersky, Kevin and Fleet, David},
  booktitle={International Conference on Learning Representations},
  volume={2024},
  pages={4793--4822},
  year={2024}
}

@inproceedings{agarwal2024policy,
  title={On-policy distillation of language models: Learning from self-generated mistakes},
  author={Agarwal, Rishabh and Vieillard, Nino and Zhou, Yongchao and Stanczyk, Piotr and Ramos Garea, Sabela and Geist, Matthieu and Bachem, Olivier},
  booktitle={International Conference on Learning Representations},
  volume={2024},
  pages={21246--21263},
  year={2024}
}

@article{mildenhall2021nerf,
  title={Nerf: Representing scenes as neural radiance fields for view synthesis},
  author={Mildenhall, Ben and Srinivasan, Pratul P and Tancik, Matthew and Barron, Jonathan T and Ramamoorthi, Ravi and Ng, Ren},
  journal={Communications of the ACM},
  volume={65},
  number={1},
  pages={99--106},
  year={2021},
  publisher={ACM New York, NY, USA}
}

@article{poole2022dreamfusion,
  title={Dreamfusion: Text-to-3d using 2d diffusion},
  author={Poole, Ben and Jain, Ajay and Barron, Jonathan T and Mildenhall, Ben},
  journal={arXiv preprint arXiv:2209.14988},
  year={2022}
}

@inproceedings{lin2023magic3d,
  title={Magic3d: High-resolution text-to-3d content creation},
  author={Lin, Chen-Hsuan and Gao, Jun and Tang, Luming and Takikawa, Towaki and Zeng, Xiaohui and Huang, Xun and Kreis, Karsten and Fidler, Sanja and Liu, Ming-Yu and Lin, Tsung-Yi},
  booktitle={Proceedings of the IEEE/CVF conference on computer vision and pattern recognition},
  pages={300--309},
  year={2023}
}

@inproceedings{sanghi2022clip,
  title={Clip-forge: Towards zero-shot text-to-shape generation},
  author={Sanghi, Aditya and Chu, Hang and Lambourne, Joseph G and Wang, Ye and Cheng, Chin-Yi and Fumero, Marco and Malekshan, Kamal Rahimi},
  booktitle={Proceedings of the IEEE/CVF conference on computer vision and pattern recognition},
  pages={18582--18592},
  year={2022}
}

@article{wu2016learning,
  title={Learning a probabilistic latent space of object shapes via 3d generative-adversarial modeling},
  author={Wu, Jiajun and Zhang, Chengkai and Xue, Tianfan and Freeman, Bill and Tenenbaum, Josh},
  journal={Advances in neural information processing systems},
  volume={29},
  year={2016}
}

@inproceedings{yan2022shapeformer,
  title={Shapeformer: Transformer-based shape completion via sparse representation},
  author={Yan, Xingguang and Lin, Liqiang and Mitra, Niloy J and Lischinski, Dani and Cohen-Or, Daniel and Huang, Hui},
  booktitle={Proceedings of the IEEE/CVF conference on computer vision and pattern recognition},
  pages={6229--6239},
  year={2022}
}

@article{yin2025shapegpt,
  title={Shapegpt: 3d shape generation with a unified multi-modal language model},
  author={Yin, Fukun and Chen, Xin and Zhang, Chi and Jiang, Biao and Zhao, Zibo and Liu, Wen and Yu, Gang and Chen, Tao},
  journal={IEEE Transactions on Multimedia},
  volume={27},
  pages={4107--4120},
  year={2025},
  publisher={IEEE}
}

@article{ho2020denoising,
  title={Denoising diffusion probabilistic models},
  author={Ho, Jonathan and Jain, Ajay and Abbeel, Pieter},
  journal={Advances in neural information processing systems},
  volume={33},
  pages={6840--6851},
  year={2020}
}

@inproceedings{rombach2022high,
  title={High-resolution image synthesis with latent diffusion models},
  author={Rombach, Robin and Blattmann, Andreas and Lorenz, Dominik and Esser, Patrick and Ommer, Bj{\"o}rn},
  booktitle={Proceedings of the IEEE/CVF conference on computer vision and pattern recognition},
  pages={10684--10695},
  year={2022}
}

@inproceedings{gu2024minillm,
  title={Minillm: Knowledge distillation of large language models},
  author={Gu, Yuxian and Dong, Li and Wei, Furu and Huang, Minlie},
  booktitle={International Conference on Learning Representations},
  volume={2024},
  pages={32694--32717},
  year={2024}
}

@article{song2023consistency,
  title={Consistency models},
  author={Song, Yang and Dhariwal, Prafulla and Chen, Mark and Sutskever, Ilya},
  journal={arXiv preprint arXiv:2303.01469},
  year={2023}
}

@article{luo2023latent,
  title={Latent consistency models: Synthesizing high-resolution images with few-step inference},
  author={Luo, Simian and Tan, Yiqin and Huang, Longbo and Li, Jian and Zhao, Hang},
  journal={arXiv preprint arXiv:2310.04378},
  year={2023}
}

@article{zhao2023michelangelo,
  title={Michelangelo: Conditional 3d shape generation based on shape-image-text aligned latent representation},
  author={Zhao, Zibo and Liu, Wen and Chen, Xin and Zeng, Xianfang and Wang, Rui and Cheng, Pei and Fu, Bin and Chen, Tao and Yu, Gang and Gao, Shenghua},
  journal={Advances in neural information processing systems},
  volume={36},
  pages={73969--73982},
  year={2023}
}

@article{li2024craftsman3d,
  title={Craftsman3d: High-fidelity mesh generation with 3d native generation and interactive geometry refiner},
  author={Li, Weiyu and Liu, Jiarui and Yan, Hongyu and Chen, Rui and Liang, Yixun and Chen, Xuelin and Tan, Ping and Long, Xiaoxiao},
  journal={arXiv preprint arXiv:2405.14979},
  year={2024}
}

@inproceedings{ye2025hi3dgen,
  title={Hi3dgen: High-fidelity 3d geometry generation from images via normal bridging},
  author={Ye, Chongjie and Wu, Yushuang and Lu, Ziteng and Chang, Jiahao and Guo, Xiaoyang and Zhou, Jiaqing and Zhao, Hao and Han, Xiaoguang},
  booktitle={Proceedings of the IEEE/CVF International Conference on Computer Vision},
  pages={01--12},
  year={2025}
}

@inproceedings{xue2023ulip,
  title={Ulip: Learning a unified representation of language, images, and point clouds for 3d understanding},
  author={Xue, Le and Gao, Mingfei and Xing, Chen and Mart{\'\i}n-Mart{\'\i}n, Roberto and Wu, Jiajun and Xiong, Caiming and Xu, Ran and Niebles, Juan Carlos and Savarese, Silvio},
  booktitle={Proceedings of the IEEE/CVF conference on computer vision and pattern recognition},
  pages={1179--1189},
  year={2023}
}

@inproceedings{zhou2024uni3d,
  title={Uni3d: Exploring unified 3d representation at scale},
  author={Zhou, Junsheng and Wang, Jinsheng and Ma, Baorui and Liu, Yu-Shen and Huang, Tiejun and Wang, Xinlong},
  booktitle={International Conference on Learning Representations},
  volume={2024},
  pages={46766--46782},
  year={2024}
}

@article{huang2023t2i,
  title={T2i-compbench: A comprehensive benchmark for open-world compositional text-to-image generation},
  author={Huang, Kaiyi and Sun, Kaiyue and Xie, Enze and Li, Zhenguo and Liu, Xihui},
  journal={Advances in Neural Information Processing Systems},
  volume={36},
  pages={78723--78747},
  year={2023}
}

@article{ghosh2023geneval,
  title={Geneval: An object-focused framework for evaluating text-to-image alignment},
  author={Ghosh, Dhruba and Hajishirzi, Hannaneh and Schmidt, Ludwig},
  journal={Advances in Neural Information Processing Systems},
  volume={36},
  pages={52132--52152},
  year={2023}
}

@article{ko2024distillm,
  title={Distillm: Towards streamlined distillation for large language models},
  author={Ko, Jongwoo and Kim, Sungnyun and Chen, Tianyi and Yun, Se-Young},
  journal={arXiv preprint arXiv:2402.03898},
  year={2024}
}

@article{wu2026direct3d,
  title={Direct3d-s2: Gigascale 3d generation made easy with spatial sparse attention},
  author={Wu, Shuang and Lin, Youtian and Zhang, Feihu and Zeng, Yifei and Yang, Yikang and Qian, Jiachen and Zhu, Siyu and Cao, Xun and Torr, Philip and Yao, Yao and others},
  journal={Advances in Neural Information Processing Systems},
  volume={38},
  pages={170778--170804},
  year={2026}
}

@article{kingma2021variational,
  title={Variational diffusion models},
  author={Kingma, Diederik and Salimans, Tim and Poole, Ben and Ho, Jonathan},
  journal={Advances in neural information processing systems},
  volume={34},
  pages={21696--21707},
  year={2021}
}

@article{gemini2025flash,
  title={Gemini 3 Flash},
  author={{Google DeepMind}},
  journal={Google AI Blog},
  year={2025}
}

@inproceedings{xiang2026native,
  title={Native and compact structured latents for 3d generation},
  author={Xiang, Jianfeng and Chen, Xiaoxue and Xu, Sicheng and Wang, Ruicheng and Lv, Zelong and Deng, Yu and Zhu, Hongyuan and Dong, Yue and Zhao, Hao and Yuan, Nicholas Jing and others},
  booktitle={Proceedings of the IEEE/CVF Conference on Computer Vision and Pattern Recognition},
  pages={14419--14429},
  year={2026}
}

\clearpage
\newpage
\appendix

\newpage
\section*{Supplementary Material}

\begin{table*}[!ht]
\centering
\resizebox{\textwidth}{!}{
\begin{tabular}{l|c|c|c}
\toprule
Hyperparameter & Stage 1a: Semi-Policy FT & Stage 1b: On-Policy DPO & Stage 2: Multi-Teacher OPD \\
\midrule
Base / Initialization & Hunyuan3D-2.1 (3B DiT) & Fine-tuned model (1a) & Fine-tuned model (1a) \\
DPO $\beta$ & 1000 & 1000 & -- \\
$\lambda_\text{dpo}$ / $\lambda_\text{distill}$ & 0.2 & 1.0 & 1.0 \\
DPO type & Semi-policy & On-policy (fully online) & -- \\
\midrule
Rollout sampler & SDE ($a$=0.3) & SDE ($a$=0.3) & ODE (deterministic) \\
Rollout NFE & 50 & 50 & 50 \\
Candidates per step & 1 (loser only) & 2 (pairs) & 1 \\
Confidence margin $\delta$ & -- & 0.167 & -- \\
Run mode & End-to-end & Iterative rollout/train(400/100) & End-to-end \\
\bottomrule
\end{tabular}
}
\caption{Hyperparameter configurations across all training stages.}
\label{tab:supp-config}
\end{table*}

\begin{table}[!t]
\centering
\resizebox{0.9\columnwidth}{!}{
\begin{tabular}{lcc}
\toprule
Dimension & Train Samples & Eval Samples \\
\midrule
Orientation & 3,661 & 133  \\
Pose & 4,083 & 111 \\
Consistency & 1,063 & 100 \\
\bottomrule
\end{tabular}
}
\caption{Dataset statistics across the three quality dimensions.}
\label{tab:supp-data}
\end{table}


\subsection{Details of the Training Pipeline}
\label{sec:supp-config}

\subsubsection{Hyperparameter Configuration}
Table~\ref{tab:supp-config} summarizes the hyperparameter configurations for all training stages. All stages are trained with 5K steps on $2\times8$ H20(90G) GPUS. The pipeline begins with semi-policy fine-tuning (stage 1a) on the offline ground-truth winners and on-policy generated losers. The resulting fine-tuned model serves as the initialization for both the domain-specialized teachers (stage 1b) and the final student (stage 2). Each teacher is trained via on-policy DPO with Gemini-3-Flash pairwise reward using an iterative rollout-then-train strategy to handle the low valid-pair ratio. In stage 2, the three frozen teachers are distilled into the student through on-policy trajectory sampling with hard task-routing and gradient accumulation across the three data types.

\subsubsection{Dataset Statistics}

Table~\ref{tab:supp-data} reports the dataset sizes and the description for each quality dimension. Specifically, in the training stage, we curate three separate preference datasets, one per dimension: an orientation set targeting canonical axis alignment, a pose set covering diverse articulated human configurations, and a consistency set emphasizing multi-view structural and textural coherence. In the evaluation stage, to isolate the model's generation ability along each aspect, we construct three dedicated test sets that mirror the three training dimensions.

\begin{figure}[!ht]
    \centering
    \includegraphics[width=\linewidth]{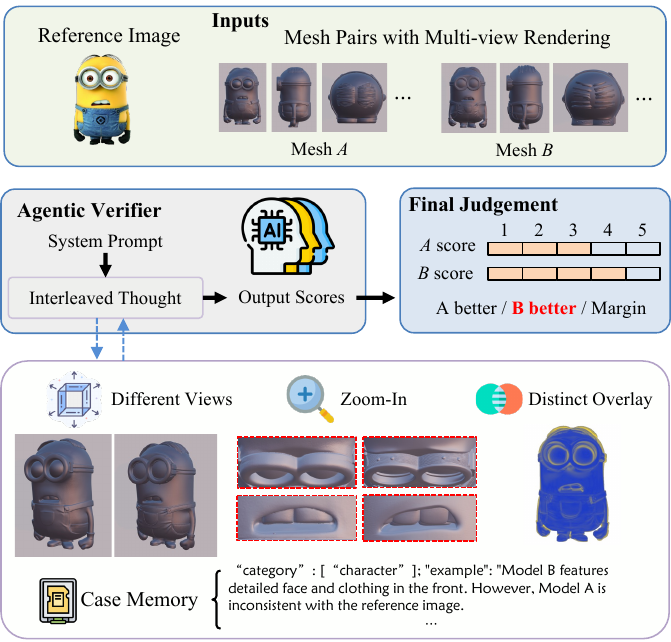}
    \caption{The detailed pipeline of our agentic verifier.}
    \label{fig:verifier}
\vspace{-1.0em}
\end{figure}

\subsection{Agentic Verifier Details}
\label{sec:supp-reward}
In this section, we introduce the details of the agentic verifier used to construct preference pairs during DPO training, depicted in Fig.~\ref{fig:verifier}. The verifier is built on the Gemini-3-Flash proprietary model and follows a structured, tool-augmented reasoning process to compare two candidate meshes generated from the same reference image. Below we describe its system prompt, tool call, and case-memory mechanism.

\subsubsection{System Prompt}

The reward agent receives a structured system prompt that defines its role, input context, dimension-specific instructions, and metrics. The key components are shown as follows.

\paragraph{Role and Input.} The agent acts as a strict 3D geometry review agent that compares two candidate meshes (model \textit{A} vs. model \textit{B}) generated from the same reference image. It receives: (1) the reference image, (2) multi-view rendered images of two meshes.

\paragraph{Dimension-Specific Instructions.} Based on the shared system prompt, we design a dedicated instruction for each of the three judge dimensions. In the orientation dimension, the agent enforces canonical axis alignment and treats perspective baking as a fatal defect. In the human-pose dimension, the agent prioritizes correct limb counts and connectivity, then joint angles and overall posture relative to the reference. In the consistency dimension, the agent checks that structures and texture inferred from one viewpoint remain plausible and coherent with the reference image. The three prompts are summarized below.

\begin{promptbox}[label=box:orient-prompt]{Orientation Prompt}
\small\ttfamily
You are a review agent focused on canonical orientation. The generated 3D model must be aligned to the standard world axes, while baking in the diagonal tilt, skewed perspective, or dynamic camera angle of the 2D reference is strictly prohibited. For example, an object with a dominant axis (e.g., a trumpet, a standing animal) must have that axis horizontally or vertically aligned in the orthographic views, not slanted. A bilaterally symmetric object (e.g., a forward-facing human) must appear macroscopically symmetric in the front view. Any retained diagonal slant or perspective baking is a fatal defect and must be scored ``Bad''.
\end{promptbox}

\begin{promptbox}[label=box:pose-prompt]{Pose Prompt}
\small\ttfamily
You are a review agent focused on human pose and body structure. Evaluate, in descending priority: Macro limb structure -- exactly two arms and two legs, each a single continuous limb, with no forking, and with limbs/head smoothly connected to the torso (no floating or severed parts). Pose accuracy -- limb angles, bend directions, and global posture must match the reference; penalize mirrored poses, collapsed poses, and residual T-pose. Overall shape and accessories -- clothing, hats, and attachments should be consistent with the reference.
\end{promptbox}

\begin{promptbox}[label=box:consist-prompt]{Consistency Prompt}
\small\ttfamily
You are a senior 3D geometry review agent specialized in geometric consistency. Your task is to evaluate how faithfully two 3D models reproduce the geometric or textural structures visible in the reference image (Image-to-3D consistency). You evaluate whether the 3D model accurately reconstructs what is shown in the reference image — silhouette match, structural element count, proportions, and geometric details. You do NOT evaluate orientation alignment or artistic style — only consistency with the reference.
\end{promptbox}

\subsubsection{Tool Interface Specifications}

The agent has access to three tools during evaluation:

\paragraph{\texttt{render\_extra\_views}} renders additional viewpoints of both models for closer inspection.
\begin{itemize}
\item \textbf{Parameters}:
  \begin{itemize}
  \item \texttt{angles}: List of viewpoints (1--6 items), each specifying \texttt{yaw} (0=Front, 90=Left, 180=Back, 270=Right), \texttt{pitch} (0=eye-level, $\pm$30=tilted, $\pm$90=top/bottom), \texttt{label} (descriptive name), and optional \texttt{render\_normal} (boolean).
  \item \texttt{reason}: Text explanation of why these views are needed.
  \end{itemize}
\item \textbf{Returns}: The rendered images at the specified viewpoints, plus corresponding difference maps.
\end{itemize}

\paragraph{\texttt{render\_zoom\_views}} renders zoomed close-up views for fine detail inspection.
\begin{itemize}
\item \textbf{Parameters}:
  \begin{itemize}
  \item \texttt{zoom\_configs}: List of zoom viewpoints (1--4 items), each specifying \texttt{yaw}, \texttt{pitch}, \texttt{label}, and optional \texttt{fov\_deg} controlling zoom level ($\sim$50$^\circ$=standard, $\sim$30$^\circ$=moderate zoom, $\sim$15$^\circ$=strong zoom, $\sim$8$^\circ$=extreme close-up).
  \item \texttt{reason}: Text explanation of what detail to inspect.
  \end{itemize}
\item \textbf{Returns}: Zoomed rendered images at the specified viewpoint and zoom level.
\end{itemize}

\paragraph{\texttt{different\_overlap}} outputs a heatmap that facilitates the agent to evaluate the difference.
\begin{itemize}
\item \textbf{Parameters}:
  \begin{itemize}
  \item \texttt{image\_index}: The index of the rendered image pairs that the agent needs to compare.
  \end{itemize}
\item \textbf{Returns}: Difference heatmap between the rendered images of the two meshes from the same viewpoint
\end{itemize}

\subsubsection{Case Memory Mechanism}
The agent maintains a case memory of previously judged hard examples to improve judgment accuracy. Before evaluating a new pair, the system retrieves similar past cases (based on object category) and includes them as few-shot examples in the context. This mechanism addresses the variance inherent in VLM-based evaluation by anchoring judgments to established precedents.

\subsection{Derivation of OPD in Flow-Matching}
\label{sec:derivation-opd}

We provide the complete derivation of reverse-KL on-policy distillation (OPD) objective in Eq.~\ref{eq:opd-loss}.  Let $\pi_{\theta}$ and $\pi_{\phi_m}$ denote the student and teacher policies. The denoising trajectory of the student policy is defined as $x_{0:N} = (x_{t_0}, \dots, x_{t_N})$, where $1 = t_0 > t_1 > \cdots > t_N = 0$. Considering the denoising process is Markovian, we decompose the whole OPD objective into stepwise KL along the student rollout:
\begin{equation}
\label{eq:supp-opd-generic}
\adjustbox{max width=0.85\linewidth}{$\displaystyle
\mathcal{L}_{\text{OPD}}(\theta)
= \mathbb{E}_{x_{0:N} \sim \pi_\theta}\!\left[
    \sum_{j=0}^{N-1}
    \mathrm{KL}\!\big(
      \pi_\theta(\cdot \mid x_{t_j}) \,\big\Vert\, \pi_{\phi_m}(\cdot \mid x_{t_j})
    \big)
\right],
$}
\end{equation}
where $\pi_\theta(\cdot \mid x_{t_j})$ and $\pi_{\phi_m}(\cdot \mid x_{t_j})$ are the student and teacher one-step transition kernels at the same state $x_{t_j}$.

For flow-matching denoising with SDE sampling, we discretize the reverse time by Euler--Maruyama with step $\Delta t_j := t_{j+1} - t_j < 0$. Let $\sigma_{t} = a\sqrt{t/(1-t)}$ be the SDE diffusion coefficient with global noise level $a$, and define $v_\theta(x_{t_j})$ and $v_{\phi_m}(x_{t_j})$ for the student and teacher velocity fields in $t_j$. The student SDE step is:
 \begin{equation}
  \label{eq:supp-sde-step}
  \resizebox{1.0\linewidth}{!}{$\displaystyle
  \begin{aligned}
  &x_{t_{j+1}} = x_{t_j} + \Big[\, v_\theta + \tfrac{\sigma_{t_j}^2}{2 t_j}\big(x_{t_j} + (1-t_j)\,v_\theta\big)\Big]\Delta t_j + \sigma_{t_j}\sqrt{-\Delta t_j}\,\varepsilon_j \\
  &= \Big(1 + \tfrac{\sigma_{t_j}^2}{2 t_j}\Delta t_j\Big) x_{t_j} + \Big(1 + \tfrac{\sigma_{t_j}^2(1-t_j)}{2 t_j}\Big) v_\theta\,\Delta t_j + \sigma_{t_j}\sqrt{-\Delta t_j}\,\varepsilon_j
  \end{aligned}
  $}
  \end{equation}
where $\varepsilon_j \sim \mathcal{N}(0, I_d)$, and $v_\theta := v_\theta(x_{t_j})$. Collecting the deterministic drift as the transition mean and the noise term as the covariance, the one-step kernel is the isotropic Gaussian:
\begin{equation}
\footnotesize
\label{eq:supp-kernel}
\pi_\theta(\cdot \mid x_{t_j}) = \mathcal{N}\!\big(\mu_\theta(x_{t_j}),\; \bar\sigma_{t_j}^2 I_d\big),
\quad \bar\sigma_{t_j}^2 := \sigma_{t_j}^2(-\Delta t_j),
\end{equation}
where the mean is defined as:
\begin{equation}
\label{eq:supp-mean}
\mu_\theta(x_{t_j}) = \Big(1 + \tfrac{\sigma_{t_j}^2}{2 t_j}\Delta t_j\Big) x_{t_j}
  + \Big(1 + \tfrac{\sigma_{t_j}^2(1-t_j)}{2 t_j}\Big) v_\theta \,\Delta t_j.
\end{equation}

The transition kernel in the teacher policy $\pi_{\phi_m}$ is defined in a similar formulation:
\begin{equation}
  \resizebox{0.85\linewidth}{!}{$\displaystyle
\begin{aligned}
\label{eq:supp-teacher-mean}
\pi_{\phi_m}(\cdot \mid x_{t_j}) &= \mathcal{N}\!\big(\mu_{\phi_m}(x_{t_j}),\; \bar\sigma_{t_j}^2 I_d\big), 
\\
\mu_{\phi_m}(x_{t_j}) = \Big(1 +& \tfrac{\sigma_{t_j}^2}{2 t_j}\Delta t_j\Big) x_{t_j}
  + \Big(1 + \tfrac{\sigma_{t_j}^2(1-t_j)}{2 t_j}\Big) v_{\phi_m} \,\Delta t_j.
\end{aligned}
$}
\end{equation}

Because the covariance $\bar\sigma_{t_j}^2 I_d$ is identical, $\pi_\theta$ and $\pi_{\phi_m}$ differ only in their means. For two $d$-dimensional Gaussians with common isotropic covariance $\sum$:
\begin{equation}
\adjustbox{max width=0.9\linewidth}{$\displaystyle
\label{eq:supp-gauss-kl}
\mathrm{KL}\!\big(\mathcal{N}(\mu_1, \textstyle{\sum})\,\Vert\,\mathcal{N}(\mu_2, \sum)\big)
= \frac{1}{2}(\mu_1-\mu_2)^{T}\textstyle{\sum^{-1}}(\mu_1-\mu_2).
$}
\end{equation}

Applying~\eqref{eq:supp-gauss-kl} to the kernels~\eqref{eq:supp-kernel} and~\eqref{eq:supp-teacher-mean} gives the per-step KL as:
\begin{equation}
\label{eq:supp-kl-mean}
\adjustbox{max width=0.85\linewidth}{$\displaystyle
\mathrm{KL}\!\big(\pi_\theta(\cdot\mid x_{t_j})\,\Vert\,\pi_{\phi_m}(\cdot\mid x_{t_j})\big)
= \frac{\|\mu_\theta(x_{t_j}) - \mu_{\phi_m}(x_{t_j})\|_2^2}{2\,\bar\sigma_{t_j}^2}.
$}
\end{equation}

Subtracting $\mu_{\phi_m}(x_{t_j})$ from $\mu_{\theta}(x_{t_j})$, the state-dependent terms cancel, and only the velocity-dependent term remains:
\begin{align}
\footnotesize
\label{eq:supp-mean-diff}
\mu_\theta&(x_{t_j}) - \mu_{\phi_m}(x_{t_j}) \notag
= \\
&\Big(1 + \tfrac{\sigma_{t_j}^2(1-t_j)}{2 t_j}\Big)\Delta t_j \,
  \big(v_\theta(x_{t_j}) - v_{\phi_m}(x_{t_j})\big).
\end{align}

Substituting~\eqref{eq:supp-mean-diff} into~\eqref{eq:supp-kl-mean}, the per-step KL becomes a weighted squared velocity difference:
\begin{equation}
  \resizebox{1.0\linewidth}{!}{$\displaystyle
\begin{aligned}
\label{eq:supp-kl-vel}
&\mathrm{KL}\!\big(\pi_\theta \,\Vert\, \pi_{\phi_m}\big)
=  w(t_j)
  \big\| v_\theta(x_{t_j}) - v_{\phi_m}(x_{t_j}) \big\|_2^2, \\
  & w(t_j)
= \frac{(-\Delta t_j)}{2\,\sigma_{t_j}^2}\Big(1 + \tfrac{\sigma_{t_j}^2(1-t_j)}{2 t_j}\Big)^2
= \frac{\Delta t_j}{2}\Big(\frac{\sigma_{t_j}(1-t_j)}{2 t_j} + \frac{1}{\sigma_{t_j}}\Big)^2.
\end{aligned}
$}
\end{equation}

Based on Eq.~\ref{eq:supp-kl-vel}, we apply the stop-gradient operator $\mathrm{SG}(\cdot)$ to the frozen teacher. The final OPD objective is the time-weighted velocity-matching loss $\mathcal{L}^{m}_{\text{distill}}(\theta)$, defined as:
\begin{equation}
\adjustbox{max width=0.85\linewidth}{$\displaystyle
\label{eq:supp-opd-final}
\mathbb{E}_{x_{0:N}\sim p_{S,\theta}}\!\left[
    \sum_{j=0}^{N-1} w(t_j)\,
    \big\| v_\theta(x_{t_j}) - \mathrm{SG}\big(v_{\phi_m}(x_{t_j})\big) \big\|_2^2
\right].
$}
\end{equation}

\paragraph{Deterministic (ODE) regime.}
Setting the noise level $a \to 0$ makes the transition deterministic: for a given $x_{t_j}$, the student and teacher each induce a single transition target. Distribution matching then reduces to pointwise matching of the predicted velocities, and the reverse-KL objective is replaced by a direct squared $L_2$ loss, which is formulated as:
\begin{equation}
\adjustbox{max width=0.85\linewidth}{$\displaystyle
\label{eq:supp-opd-ode}
 \mathbb{E}_{x_{0:N}\sim p_{S,\theta}}\!\left[
    \frac{1}{N}\sum_{j=0}^{N-1}
    \big\| v_\theta(x_{t_j}) - \mathrm{SG}\big(v_{\phi_m}(x_{t_j})\big) \big\|_2^2
\right],
$}
\end{equation}
which is the ODE instantiation used in our OPD stage. This closed-form, zero-variance objective is the flow-matching analogue of the per-step KL distillation used for autoregressive models, replacing high-variance policy gradients with direct backpropagation through the student velocity field.

\subsection{Additional Experiments}
\label{sec:additional-exp}

\subsubsection{Impact of the CFG Weights}

We further study the effect of the CFG weight $w$ in both the training and the inference stages, as reported in Table~\ref{tab:ablation-cfg-perdim} and Table~\ref{tab:ablation-cfg}. During the OPD stage, we find that using the velocity field predicted under the default conditional signal (i.e., $w=1.0$, without classifier-free amplification) as the distillation target yields the best student, indicating that the raw conditional guidance already provides a sufficiently informative supervision signal. At inference time, the model performs best under the default CFG configuration used in Hunyuan3D-2.1 ($w=3.0$), which balances conditional fidelity and generation quality.

\begin{table}[!ht]
\centering
\resizebox{\columnwidth}{!}{
\begin{tabular}{lccc}
\toprule
\multirow{2}{*}{CFG Scale $w$} & \multicolumn{3}{c}{A-wins/B-wins} \\
 & Orient-Eval & Pose-Eval & Consist-Eval \\
\midrule
\rowcolor{ourgray} $w=1.0$ & \textbf{5.1/18.8} & \textbf{0.0/93.5} & \textbf{10.0/81.0}  \\
$w=3.0$ & 7.1/17.6 & 1.0/91.9 & 15.7/77.5 \\
$w=6.0$ & 7.1/17.9 & 0.0/88.7	& 14.0/75.3 \\
\bottomrule
\end{tabular}
}
\caption{Per-dimension ablation on the CFG scale $w$ used for the teacher during the OPD stage. Best results are in \textbf{bold}.}
\label{tab:ablation-cfg-perdim}
\end{table}

\begin{table}[!ht]
\centering
\resizebox{\columnwidth}{!}{
\begin{tabular}{lcccc}
\toprule
CFG Scale $w$ & A-wins $\downarrow$ & B-wins & ULIP-I & Uni3D-I \\
\midrule
$w=1.0$ & 32.3 & 38.5 & 0.133 & 0.314 \\
\rowcolor{ourgray} $w=3.0$ & \textbf{4.9} & \textbf{61.0} & \textbf{0.138} & \textbf{0.321} \\
$w=6.0$ & 5.2 & 60.6 & 0.137 & 0.320 \\
\bottomrule
\end{tabular}
}
\caption{Ablation on the CFG scale $w$ during inference after OPD. Best results are in \textbf{bold}.}
\label{tab:ablation-cfg}
\end{table}

\begin{figure*}[t]
    \centering
    \includegraphics[width=\textwidth]{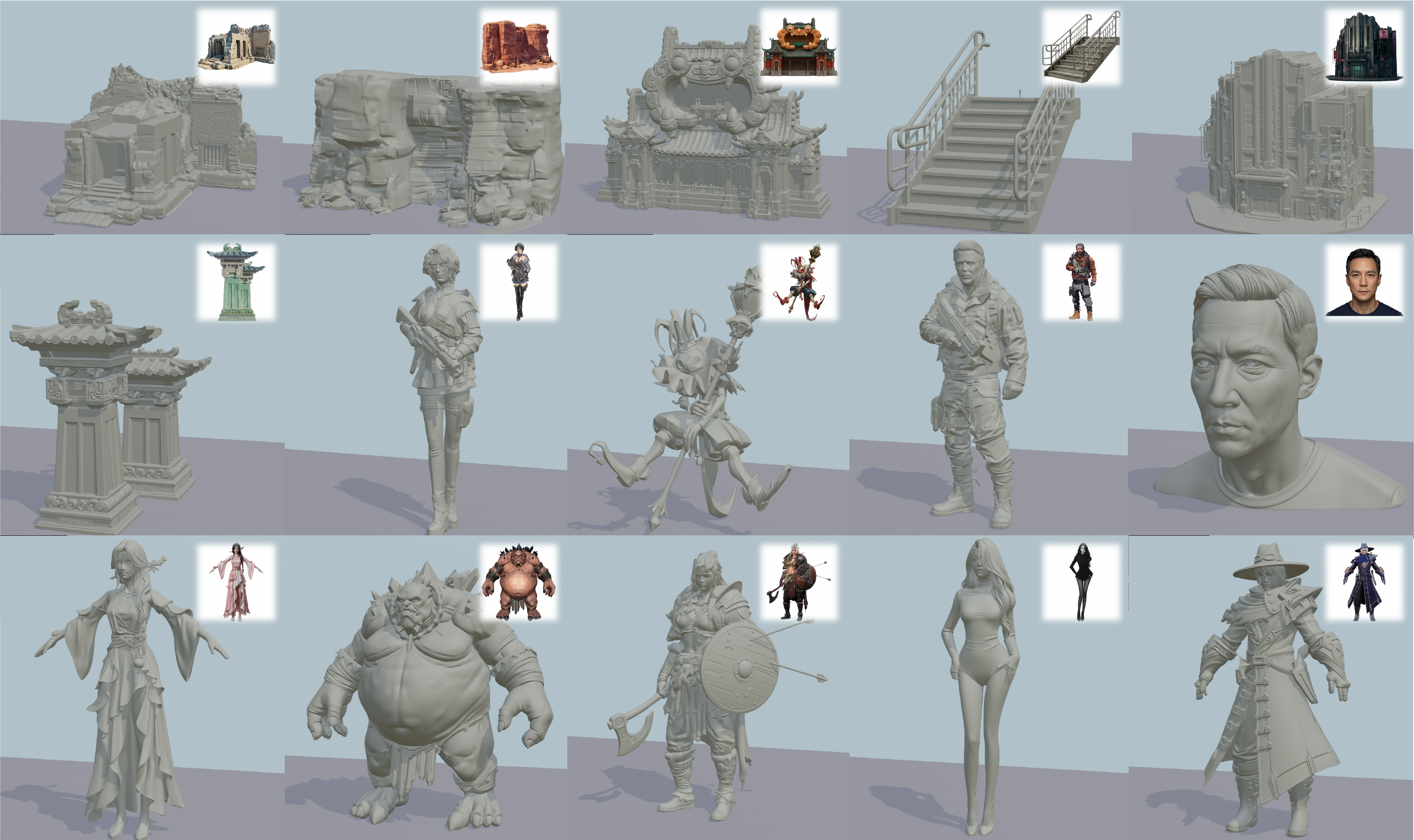}
    \caption{The quality visualization of the refinement in our \ours}
    \label{fig:qualitative}
\vspace{-1.0em}
\end{figure*}

\subsubsection{More Quality Visualization}

In this section, we provide additional quality visualizations, as shown in Fig.~\ref{fig:qualitative}. These cases are based on meshes generated by our \ours{} and further post-processed with LATTICE for high-resolution refinement.

\end{document}